\documentclass[letterpaper]{article} % DO NOT CHANGE THIS
\usepackage{aaai25}  % DO NOT CHANGE THIS
\usepackage{times}  % DO NOT CHANGE THIS
\usepackage{helvet}  % DO NOT CHANGE THIS
\usepackage{courier}  % DO NOT CHANGE THIS
\usepackage[hyphens]{url}  % DO NOT CHANGE THIS
\usepackage{graphicx} % DO NOT CHANGE THIS
\usepackage{natbib}  % DO NOT CHANGE THIS AND DO NOT ADD ANY OPTIONS TO IT
\usepackage{caption} % DO NOT CHANGE THIS AND DO NOT ADD ANY OPTIONS TO IT
\usepackage{algorithm}
\usepackage{algorithmic}

\usepackage{booktabs}
\usepackage{multirow}
\usepackage{multicol}
\usepackage{amsmath}
\usepackage{makecell}
\usepackage[dvipsnames]{xcolor}
\usepackage{amssymb}
\usepackage{marvosym}
\usepackage{newfloat}
\usepackage{listings}
\DeclareCaptionStyle{ruled}{labelfont=normalfont,labelsep=colon,strut=off} % DO NOT CHANGE THIS
\floatstyle{ruled}
\newfloat{listing}{tb}{lst}{}
\floatname{listing}{Listing}
\title{Implicit Word Reordering with Knowledge Distillation for Cross-Lingual Dependency Parsing}
\author{
    Zhuoran Li,
    Chunming Hu,
    Junfan Chen,
    Zhijun Chen,
    Richong Zhang\thanks{Corresponding author.}
}
\affiliations{
    SKLSDE, Beihang University, Beijing, China\\
    \{lizhuoranget, hucm, zhijunchen\}@buaa.edu.cn,
    \{chenjf, zhangrc\}@act.buaa.edu.cn
}

\begin{document}

\maketitle

\begin{abstract}
Word order difference between source and target languages is a major obstacle to cross-lingual transfer, especially in the dependency parsing task. Current works are mostly based on order-agnostic models or word reordering to mitigate this problem. However, such methods either do not leverage grammatical information naturally contained in word order or are computationally expensive as the permutation space grows exponentially with the sentence length. Moreover, the reordered source sentence with an unnatural word order may be a form of noising that harms the model learning. To this end, we propose an Implicit Word Reordering framework with Knowledge Distillation (\textbf{IWR-KD}). This framework is inspired by that deep networks are good at learning feature linearization corresponding to meaningful data transformation, e.g. word reordering. To realize this idea, we introduce a knowledge distillation framework composed of a word-reordering teacher model and a dependency parsing student model. We verify our proposed method on Universal Dependency Treebanks across 31 different languages and show it outperforms a series of competitors, together with experimental analysis to illustrate how our method works towards training a robust parser.
\end{abstract}

\section{Introduction}

Dependency parsing is a fundamental task which aims to extract the low-level grammatical relationships between words in a sentence~\cite{10.5555/1237357, nivre-2008-algorithms, kiperwasser-goldberg-2016-simple}, such as subject-verb relationships. Recently, cross-lingual dependency parsing has attracted considerable attention from academic and industrial communities, for which a parser is trained on a source language and directly applied in the target language of interest. Multilingual pre-trained language models (mPLMs) demonstrate exceptional performance in cross-lingual dependency parsing~\cite{devlin-etal-2019-bert, conneau-etal-2020-unsupervised}. However, these mPLMs inevitably encode word order features to model contextual representations, referred to as \textit{order-sensitive}~\cite{ahmad-etal-2019-difficulties, Liu_Winata_Cahyawijaya_Madotto_Lin_Fung_2021}. Since {word order is inherently different across languages}, there is a risk of over-fitting into the word order of the source language that could hurt the performance in the target languages.

\begin{figure}[!t]
\centering
\begin{tabular}{c}
    \includegraphics[width=0.98\linewidth]{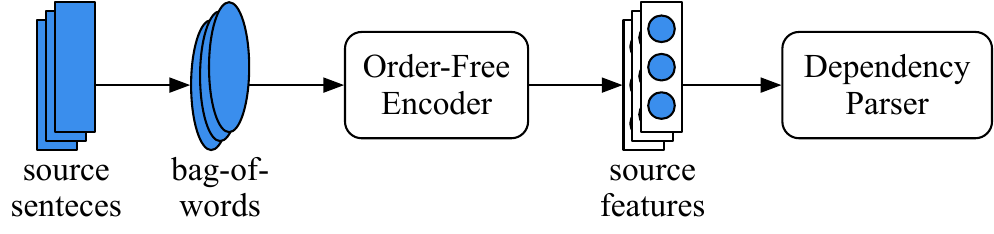}    \\
    (a) Order-agnostic \\   
    \includegraphics[width=0.98\linewidth]{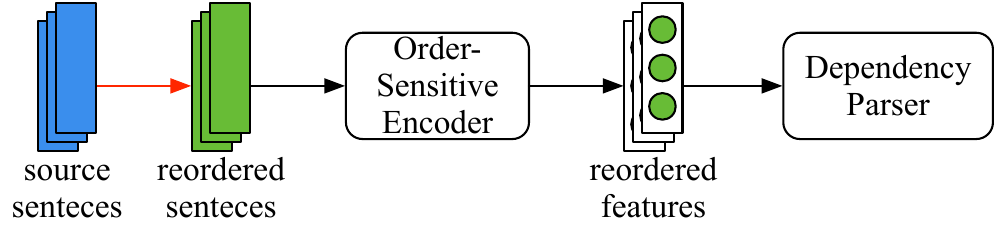}    \\
    (b) Explicit word reordering \\ \includegraphics[width=0.98\linewidth]{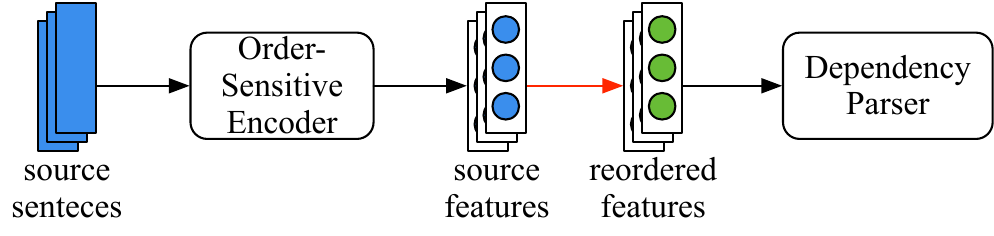}  \\ 
    (c) Implicit word reordering (ours)
\end{tabular}
\caption{Comparison between different methods for Word Ordering Difference in cross-lingual dependency parsing. (a) Removing word order information. (b) Permuting the words in a source sentence to resemble the word order of a target language. (c) Our method adapts the word order in the feature space. Red arrows indicate reordering steps.}
\label{fig:intro}
\end{figure}

Therefore, a lot of works pay attention to the word order difference problem in cross-lingual dependency parsing. The methods mainly fall into two categories: First, as shown in Figure \ref{fig:intro} (a), order-agnostic based methods, which utilize order-free encoders, e.g. Self-Attention~\cite{ahmad-etal-2019-difficulties}, Frozen Position Embedding~\cite{Liu_Winata_Cahyawijaya_Madotto_Lin_Fung_2021}, Bag-of-Words~\cite{ji-etal-2021-word}, to maintain robust performance to the change of word order. Second, as shown in Figure \ref{fig:intro} (b), word reordering-based methods~\cite{rasooli-collins-2019-low, liu-etal-2020-cross-lingual-dependency, arviv-etal-2023-improving}, which generally first generate a large set of new sentences syntactically similar to the target language via rearranging words in the source sentence, and then select generated high-quality sentences to train a target-language dependency parser. Since these word reordering methods essentially rearrange words in the source language, we call this category of methods as \textit{explicit word reordering} (EWR).

However, there are limitations to the previous methods. The order-agnostic methods could weaken its representation capability due to the lack of word order information, resulting in a drop in dependency parsing. The EWR methods can be computationally expensive, as the permutation space grows exponentially with the sentence length. Moreover, the explicit word rearrangement may introduce linguistic adversity, i.e. unnatural word order in the source language, and therefore can be seen as a form of noising that harms the model learning~\cite{wei-etal-2021-shot, arviv-etal-2023-improving}.

To address the limitations, we explore a new attempt, {\em implicit word reordering} and propose an Implicit Word Reordering method with Knowledge Distillation (IWR-KD) for cross-lingual dependency parsing. Motivated by that deep learning are surprisingly good at features linearization~\cite{pmlr-v28-bengio13, DBLP:conf/cvpr/UpchurchGPPSBW17, NIPS2019_9426}, IWR-KD learns to implicitly adapt the word order relationship in the word representation space rather than really permuting the words given a sentence. Figure \ref{fig:intro} (c) shows the differences between the proposed IWR-KD method and the typical order-agnostic and EWR-based methods.

Specifically, we train a word order model using the target-language part-of-speech (POS) data\footnote{In this paper, we assume that our corpora are annotated with gold POS tags, even in target languages that lack gold training trees. While this is an idealized setting, it is sufficient to achieve good results. This approach has often been used in studies on unsupervised and cross-lingual transfer. Our ablation study explores a potential alternative that does not rely on gold tags.}, which is much easier to annotate than the parsing tree and often reflects the syntactic structure of a language~\cite{liu-etal-2020-cross-lingual-dependency}. This trained target-language word order prediction model is then used as a teacher model to decide the new order between two POS tags. Finally, we train a student model to mimic the order prediction of the teacher and the dependency parsing using the labelled source language training data. Once the student model is ready, it can not only on-the-fly generate representations that correspond to the target-language word order relationship, but also preserve the essential linguistic structure of the original input, thus avoiding the mentioned limitations of previous works. 

In summary, the contributions of this paper are as follows:
\begin{itemize}
    \item We propose an implicit word reordering method for cross-lingual dependency parsing, addressing the limitations of previous approaches related to word order representations and high computational costs.
    \item We introduce a word reordering teacher model to effectively incorporate the target word order knowledge into the dependency parsing student model. 
    \item We conduct extensive experiments demonstrating that IWR-KD outperforms several competing baselines on Universal Dependency Treebanks across 31 languages. %Ablation studies also show IWR is critical.
\end{itemize}

\section{Preliminaries}
In this section, we first briefly review the cross-lingual dependency parsing task, backbone and prior word reordering work. These are the foundations applied to our approach.

\paragraph{Task Description}

\begin{figure}[!t]
\centering
\begin{tabular}{c}
    \includegraphics[width=0.8\linewidth]{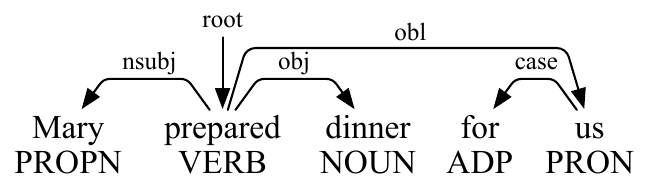}    \\
    (a) Original English sentence. \\
     \includegraphics[width=0.8\linewidth]{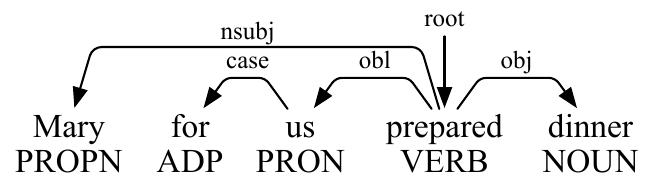}  \\ 
    (b) Explicit Estonian-specific reordered sentence.
\end{tabular}
\caption{An example of an English sentence that is explicitly reordered to resemble to the Estonian syntactic order.}
\label{fig:exp}
\end{figure}

Dependency parsing is the task of creating the dependency tree for an input sentence, which is a directed graph and defines the grammatical relation between dependent words (e.g. \textit{Mary}) and their heads (e.g. \textit{prepared}), as shown in Figure \ref{fig:exp} (a). The goal of cross-lingual dependency parsing is to train a parser on the source language and perform well on an unseen target language. Throughout this work, we systematically study the transferability of the proposed implicit word reordering between source and target languages with different word order distances. 

\begin{figure*}[!t]
\centering
\begin{tabular}{c}
    \includegraphics[width=0.99\linewidth]{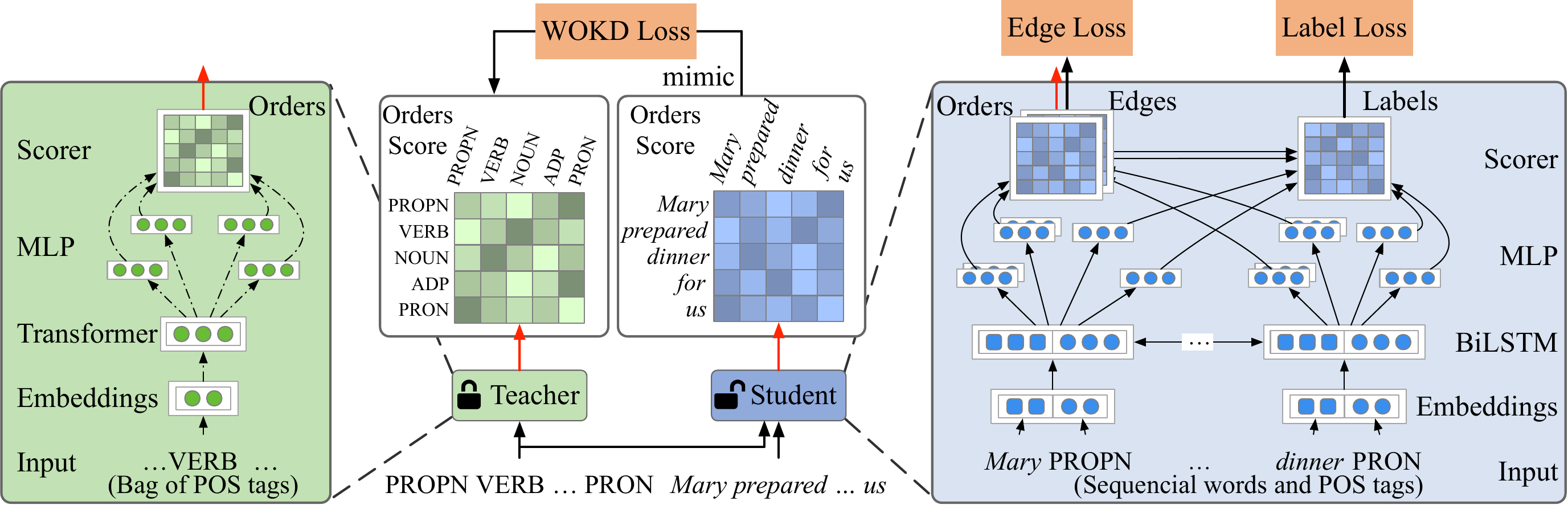}
\end{tabular}
\caption{An overview of our IWR-KD: (i) Word Reordering Teacher decides the new direction between a dependent word and its head. (ii) Dependency Parsing Student is supervised by the teacher and the gold dependency parsing labels simultaneously. }
\label{fig:overall}
\end{figure*}

\paragraph{Backbone: Biaffine Dependency Parser}
Following~\cite{ahmad-etal-2019-difficulties, wu-dredze-2019-beto, arviv-etal-2023-improving}, we adopt the graph-based bi-affine dependency parsing model \cite{dozat-manning-2018-simpler} as the backbone of our parsers, which is composed of four layers, i.e., embedding layer, Bi-LSTM layer, MLP layer and scorer layer, as shown in the right side of Figure \ref{fig:overall}. Following~\cite{ahmad-etal-2019-difficulties, Liu_Winata_Cahyawijaya_Madotto_Lin_Fung_2021}, all the parsers take words as well as their gold part-of-speech (POS) tags as input. Formally, given the input sequence $s$ with $L$ words $\{w_1, w_2,..., w_L\}$ and its POS tags $\{p_1, p_2,..., p_L\}$, the embedding layer creates a sequence of input embeddings $e_{1:L}$ in which each $e_i$ is a concatenation of its word embedding ($e_{w_i}$) and POS embedding ($e_{p_i}$). The POS embedding is trained from scratch, while the word embedding is initialized with the pre-trained multilingual language model, such as mBERT~\cite{devlin-etal-2019-bert}. 

\begin{equation}
\label{eq:embed}
    e_{i}= e_{w_i} \oplus e_{p_i}
\end{equation}
To further introduce the contextual information, we then encode each input embedding by a multilayer bidirectional LSTM:
\begin{equation}
\label{eq:bilstm}
    r_i = {\rm BiLSTM}(e_{1:L}, i)
\end{equation}

Two dimension-reducing MLPs are then used to specialise each recurrent representation $r_i$ into head-word and dependent-word representations for both the edge and the label prediction. 
\begin{equation}
\label{eq:hidden_edge_head}
    h^{\rm edge-head}_i = {\rm MLP}^{\rm edge-head}(r_i)
\end{equation}
\begin{equation}
\label{eq:hidden_edge_dep}
    h^{\rm edge-dep}_i = {\rm MLP}^{\rm edge-dep}(r_i)
\end{equation}
\begin{equation}
\label{eq:hidden_label_head}
    h^{\rm label-head}_i = {\rm MLP}^{\rm label-head}(r_i)
\end{equation}
\begin{equation}
\label{eq:hidden_label_dep}
    h^{\rm label-dep}_i = {\rm MLP}^{\rm label-dep}(r_i)
\end{equation}
Following~\cite{dozat-manning-2018-simpler}, we then use biaffine classifiers to compute the edge attention scores and label attention scores. These scores can be decoded into a graph by keeping only edges that received a positive score. 
\begin{equation}
\label{eq:score_edge}
    s^{\rm edge}_{i \to j} = h^{\rm edge - dep}_{j} U^{\rm edge} h^{{\rm edge-head}(\mathrm{T})}_{i} + b^{\rm edge}
\end{equation}
\begin{equation}
\label{eq:score_label}
    s^{\rm label}_{i \to j} = h^{\rm label - dep}_{j} U^{\rm label} h^{{\rm label-head}(\mathrm{T})}_{i} + b^{\rm label}
\end{equation}
\begin{equation}
\label{eq:y_edge}
    \hat y^{\rm edge}_{i \to j} = \{s_{i \to j} \geq 0 \}
\end{equation}
\begin{equation}
\label{eq:y_label}
    \hat y^{\rm label}_{i \to j} = \mathop{\arg\max} s^{\rm label}_{i \to j}
\end{equation}
where $U$ and $b$ are linear transformation and bias term, respectively. For each position pair $i, j$, a binary cross-entropy loss is used for the existence of edge $i \to j$, and a cross-entropy loss is used for the labels of gold edges.
\begin{equation}
\label{eq:loss_edge}
    \mathcal{L}^{\rm edge} = {\rm BCELoss}(y^{\rm edge}_{i \to j}, \hat y^{\rm edge}_{i \to j})
\end{equation}
\begin{equation}
\label{eq:loss_label}
    \mathcal{L}^{\rm label} = {\rm CELoss}(y^{\rm label}_{i \to j}, \hat y^{\rm label}_{i \to j}) 
\end{equation}
The training object for parsing is the summing of above two losses.
\begin{equation}
\label{eq:loss_parse}
    \mathcal{L} = \mathcal{L}^{\rm edge} + \lambda_1 \mathcal{L}^{\rm label}
\end{equation}

\paragraph{Prior Explicit Word Reordering} 
Given a sentence $s=\{w_1, w_2,..., w_L\}$ in the source language, prior explicit word reordering (EWR) works~\cite{Liu_Winata_Cahyawijaya_Madotto_Lin_Fung_2021, arviv-etal-2023-improving, rasooli-collins-2019-low} aim to permute the words in it to syntactically more similar to the order of the target language. Then the reordered sentences, denoted as $s'=\{w'_1, w'_2,..., w'_L\}$, as shown in Figure \ref{fig:exp} (b), used to train the cross-lingual dependency parser.

\section{Method: Implicit Word Reordering}

To mitigate the shortcomings of the existing EWR approach, we propose an \textbf{I}mplicit \textbf{W}ord \textbf{R}eordering algorithm via \textbf{K}nowledge \textbf{D}istillation \textbf{(IWR-KD)}. Unlike conventional EWR where words really are rearranged, our IWR-KD adapts word order relations in the feature space and integrates this procedure into the training of cross-lingual dependency parser. 

The overall structure of the proposed IWR-KD is shown in Figure \ref{fig:overall}, which consists of two important components: a \textit{word reordering teacher} aims to decide the new word order given the source input, and a \textit{dependency parsing student} not only learns dependency parsing from the labelled source input but also mimics target word order predictions of the teacher.
 
\subsection{Word Reordering Teacher Training}

The goal of the word reordering teacher model is to decide the word order of each dependency according to the POS tags of the dependent words and their heads. Given the target language sentence $\{w_1, w_2,..., w_L\}$ and their POS tags $\{p_1, p_2,..., p_L\}$, the following steps are applied to train the word reordering teacher: 
\begin{enumerate}
    \item \textbf{Head-Words Finding.} For each word $w_i$ in the sentence, we predict its head word $w_j$ using a pre-trained parser in the source language. 
    \item \textbf{Training Instances Extraction.} We create a training instance for each pairwise POS tags  $<p_i, p_j>$ of dependent-head words $<w_i, w_j>$. The word order label $y^{order}_{i \to j} \in \{0, 1\}$ is made based on their positions in the original sentence, where $y^{order}_{i \to j}=0$ indicates the POS tag $p_i$ is on the left of its head $p_j$, while $y^{order}_{i \to j}=1$ indicates $p_i$ is on the right of $p_j$ in the original sentence. 
    \item \textbf{Teacher Training.} As shown in the left-side in Figure \ref{fig:overall}, the teacher network consists of four layers. Given a sequence of POS tag $\{p_1, p_2,..., p_L\}$, we first obtain its POS embeddings $\{e_{p_1}, e_{p_2},..., e_{p_L}\}$. We only use the universal POS tags to discard the word form discrepancy. We then introduce contextual information by an order-free Transformer layer, which could improve cross-lingual generalization compared with the BiLSTM layer.  
    \begin{equation}
    \label{eq:transformer}
        z_{p_i}= {\rm Transformer}(e_{p_i})
    \end{equation}
    We obtain the head and dependent representation for each $z_{p_i}$ through two dimension-reducing MLPs.
    \begin{equation}
    \label{eq:hidden_order_head}
        h^{\rm order-head}_i = {\rm MLP}^{\rm order-head}(z_{p_i})
    \end{equation}
    \begin{equation}
    \label{eq:hidden_order_dep}
        h^{\rm order-dep}_i = {\rm MLP}^{\rm order-dep}(z_{p_i})
    \end{equation}
We compute word order for the given pairwise POS tags having head-dependent edges. A binary cross-entropy loss is used for the word order learning.
\begin{equation}
\label{eq:order_tea}
\hat s^{\rm order-tea}_{i \to j}= h^{\rm order - dep}_{j} U^{\rm order} h^{{\rm order-head}(\mathrm{T})}_{i} + b^{\rm order}
\end{equation}
\begin{equation}
\label{eq:y_order}
    \hat y^{\rm order-tea}_{i \to j} = \mathop{\arg\max} s^{\rm order-tea}_{i \to j}
\end{equation}
\end{enumerate}

\begin{equation}
\label{eq:loss_edge}
    \mathcal{L}^{\rm order} = {\rm BCELoss}(y^{\rm order}_{i \to j}, \hat y^{\rm order-tea}_{i \to j})
\end{equation}
After the teacher has well-trained in the target language, we take the teacher as guidance to train the dependency parsing student.

\subsection{Dependency Parsing Student Learning}
The target of the student model is to parse a sentence based on the transformed features corresponding to the reordered source sentence. As illustrated in the right side of Figure \ref{fig:overall}, we implement the student model by adding a word order learning network based on the backbone. The student model is trained to mimic the prediction probability distribution of word order generated by the teacher model on the POS tags in the target language. This process aims to transfer word order knowledge from the teacher model to the student model while allowing the student model to leverage language-specific word ordering knowledge available in the target-language POS tags.

Here, the student model consists of four layers. Specifically, given a source input sentence $s$ with $L$ words $\{w_1, w_2,..., w_L\}$ and its POS (part-of-speech) tags $\{p_1, p_2,..., p_L\}$. We first obtain its representation sequence $\{r_1, r_2,..., r_L\}$ through the embedding layer and BiLSTM layer, as shown in Equation \ref{eq:embed}-\ref{eq:bilstm}.

We then use Equation \ref{eq:hidden_edge_head}-\ref{eq:hidden_label_dep} and \ref{eq:hidden_order_head}-\ref{eq:hidden_order_dep} to specialize each recurrent representation $r_i$ into head-word $h_{1:L}^{head}$ and dependent-word $h_{1:L}^{dep}$ representations using two dimension-reduction MLPs for edge, label and word order prediction, respectively.
In addition to calculating the edge prediction and label prediction according to the Equation \ref{eq:score_edge}-\ref{eq:y_label}, we also calculate the word order prediction as follows:
\begin{equation}
\label{eq:order_stu}
s^{\rm order-stu}_{i \to j} = h^{\rm order - dep}_{j} U^{\rm order} h^{{\rm order-head}(\mathrm{T})}_{i} + b^{\rm order}
\end{equation}
\begin{equation}
\label{eq:y_order}
    \hat y^{\rm order-stu}_{i \to j} = \mathop{\arg\max} s^{\rm order-stu}_{i \to j}
\end{equation}
The student model learns dependency parsing by the following two losses:
\begin{equation}
\label{eq:loss_edge}
    \mathcal{L}^{\rm edge} = {\rm BCELoss}(y^{\rm edge}_{i \to j}, \hat y^{\rm edge-stu}_{i \to j})
\end{equation}
\begin{equation}
\label{eq:loss_label}
    \mathcal{L}^{\rm label} = {\rm CELoss}(y^{\rm label}_{i \to j}, \hat y^{\rm label-stu}_{i \to j}) 
\end{equation}
The word order knowledge distillation learning loss is formulated as the mean squared error loss:
\begin{equation}
\label{eq:loss_wokd}
    \mathcal{L}^{\rm order} = {\rm MSELoss}(\hat y^{\rm order-tea}_{i \to j}, \hat y^{\rm order-stu}_{i \to j})
\end{equation}
And the whole student training loss is the summation of three losses: 
\begin{equation}
\label{eq:loss_stu}
    \mathcal{L} = \mathcal{L}^{\rm edge} + \lambda_1 \mathcal{L}^{\rm label}+ \lambda_2 \mathcal{L}^{\rm order}
\end{equation}

\section{Experiments}
In this section, we conduct extensive experiments on 31 languages across a broad spectrum of language families to validate the effectiveness and reasonableness of our proposed IWR-KD method for cross-lingual dependency parsing.

\subsection{Setup}
\paragraph{Datasets} 
Following the setup of~\cite{ahmad-etal-2019-difficulties}, we conduct experiments on Universal Dependencies (UD) Treebanks (v2.14)~\cite{ud2.14-11234/1-5502}, in which 31 different languages are selected for evaluation. In our main experiments, we take English as the source language and 30 other languages as the target ones. We only use the source language for both training and hyper-parameter tuning. 

\paragraph{Performance Metric} The evaluation metrics are unlabeled attachment score (UAS) and labeled attachment score (LAS). Each experiment is conducted in three runs with different random seeds and the average scores are reported.

\begin{table*}[t]
    \centering
        \begin{tabular}{c | c || cccccc}
        \toprule
            \textbf{Lang.} & \textbf{Dist.} & \textbf{SelfAttn} & \textbf{mBERT} & \textbf{Frozen PE} & \textbf{Subtree-EWR} & \textbf{WOL} & \textbf{IWR-KD (ours)}\\
\midrule
en & 0.00 & 90.4/88.4 & 92.4/90.3 & 92.2/90.3 $\downarrow$ & 92.4/90.3 & \underline{93.0}/\underline{91.5} & 92.5/90.5 \\
\midrule
no & 0.06 & 80.8/72.8 & 86.7/77.4 & 87.0/78.0 & 84.5/76.6 $\downarrow$ & \underline{88.3}/\underline{78.8} & 87.9/\underline{78.8} \\
sv & 0.07 & 81.0/73.2 & 84.3/76.7 & 84.2/76.4 $\downarrow$ & 81.8/74.8 $\downarrow$ & 84.9/77.3 & \underline{85.3}/\underline{78.1} \\
fr & 0.09 & 77.9/72.8 & 83.4/71.7 & 83.4/72.0 & 84.7/\underline{73.5} & 84.4/72.5 & \underline{85.2}/73.4 \\
pt & 0.09 & 76.6/67.8 & 81.2/73.0 & 80.4/72.4 $\downarrow$ & 81.4/73.0 & 82.6/74.6 & \underline{83.1}/\underline{75.4} \\
da & 0.10 & 76.6/67.9 & 81.7/72.6 & 81.3/72.8 $\downarrow$ & 80.0/71.6 $\downarrow$ & 82.7/73.2 & \underline{83.0}/\underline{73.8} \\
es & 0.12 & 74.5/66.4 & 80.5/71.1 & 79.1/70.1 $\downarrow$ & 79.8/70.3 $\downarrow$ & 81.6/72.1 & \underline{82.3}/\underline{72.7} \\
it & 0.12 & 80.8/75.8 & 84.3/77.7 & 83.6/77.2 $\downarrow$ & 84.9/78.9 & 85.6/\underline{79.5} & \underline{85.9}/\underline{79.5} \\
ca & 0.13 & 73.8/65.1 & 80.5/70.4 & 79.1/69.4 $\downarrow$ & 81.6/\underline{72.1} & 81.5/71.5 & \underline{82.3}/71.9 \\
hr & 0.13 & 61.9/52.9 & 78.0/64.4 & 77.7/64.2 $\downarrow$ & 77.7/\underline{65.3} & 77.0/63.1 $\downarrow$ & \underline{78.2}/65.0 \\
pl & 0.13 & 74.6/62.2 & 88.4/75.7 & 87.2/75.4 $\downarrow$ & 88.3/\underline{76.8} & \underline{88.8}/75.9 & \underline{88.8}/76.6 \\
sl & 0.13 & 68.2/56.5 & 79.2/66.2 & 77.5/64.9 $\downarrow$ & 77.1/64.4 $\downarrow$ & 79.2/65.0 $\downarrow$ & \underline{79.7}/\underline{67.2} \\
uk & 0.13 & 60.1/52.3 & 73.9/63.0 & 75.8/64.5 & 73.8/62.3 $\downarrow$ & 75.8/63.8 & \underline{77.1}/\underline{65.4} \\
bg & 0.14 & 79.4/68.2 & 85.3/73.9 & 85.6/74.4 & 76.8/66.7 $\downarrow$ & \underline{87.0}/\underline{74.8} & 86.3/74.4 \\
cs & 0.14 & 63.1/53.8 & 78.3/64.2 & 77.7/64.0 $\downarrow$ & 78.4/\underline{65.0} & 77.5/63.0 $\downarrow$ & \underline{78.5}/64.5 \\
de & 0.14 & 71.3/61.6 & 77.9/69.3 & 78.7/70.3 & \underline{80.8}/\underline{73.0} & 78.7/68.6 & 79.9/70.4 \\
he & 0.14 & 55.3/48.0 & 68.0/53.8 & 68.3/54.2 & 69.7/\underline{55.1} & 68.3/54.5 & \underline{70.6}/55.0 \\
nl & 0.14 & 68.6/60.3 & 79.7/71.3 & 79.4/71.2 $\downarrow$ & 79.8/71.9 & 79.3/70.6 $\downarrow$ & \underline{80.7}/\underline{72.2} \\
ru & 0.14 & 60.6/51.6 & 77.3/65.8 & 76.0/64.9 $\downarrow$ & 75.6/65.6 $\downarrow$ & 77.3/65.9 & \underline{77.8}/\underline{66.1} \\
ro & 0.15 & 65.1/54.1 & 78.2/64.6 & 78.5/66.0 & 79.1/\underline{66.0} & \underline{79.2}/65.2 & 76.1/63.2 $\downarrow$ \\
id & 0.17 & 49.2/43.5 & 60.4/48.9 & 62.1/50.3 & \underline{70.5}/\underline{58.4} & 60.3/49.1 & 62.2/50.7 \\
sk & 0.17 & 66.7/58.2 & 82.9/69.4 & 83.4/70.9 & 82.4/69.8 $\downarrow$ & 81.8/67.8 $\downarrow$ & \underline{84.5}/\underline{70.6} \\
lv & 0.18 & 70.8/49.3 & 78.2/56.9 & 78.0/56.9 $\downarrow$ & 72.3/54.1 $\downarrow$ & 78.8/56.9 & \underline{79.2}/\underline{57.8} \\
et & 0.20 & 65.7/44.9 & 73.7/52.2 & 73.8/52.4 & 71.0/51.7 $\downarrow$ & 74.3/52.2 & \underline{74.7}/\underline{53.0} \\
fi & 0.20 & 66.3/48.7 & 74.5/53.7 & 75.4/54.9 & 71.1/53.5 $\downarrow$ & \underline{76.1}/\underline{55.5} & 75.9/55.4 \\
zh & 0.23 & 42.5/25.1 & 55.7/37.5 & 56.0/37.6 & \underline{60.0}/\underline{40.7} & 57.2/37.8 & 55.9/37.8 \\
ar & 0.26 & 38.1/28.0 & 44.8/33.1 & 47.1/35.5 & \underline{54.1}/\underline{41.1} & 46.5/34.9 & 48.5/36.2 \\
la & 0.28 & 48.0/35.2 & 57.8/41.4 & 57.0/40.6 $\downarrow$ & \underline{57.9}/\underline{42.5} & 57.0/40.9 $\downarrow$ & 56.6/40.8 $\downarrow$ \\
ko & 0.33 & 34.5/16.4 & 44.8/25.1 & 45.0/25.1 & \underline{47.9}/\underline{29.7} & 43.8/23.1 $\downarrow$ & 47.7/26.6 \\
hi & 0.40 & 35.5/26.5 & 41.4/26.8 & 42.5/28.0 & 41.0/27.0 $\downarrow$ & 40.5/27.1 $\downarrow$ & \underline{43.3}/\underline{28.2} \\
ja & 0.49 & 28.2/\underline{20.9} & 25.2/16.6 & 27.1/17.8 & 26.4/18.4 & 25.5/16.6 & \underline{28.8}/20.6 \\
\midrule
\makecell*[l]{AVG \\ \footnotesize{$\pm$STD}} & 0.17 & \makecell*[c]{64.1/53.8 \\ \footnotesize{/}} & \makecell*[c]{72.9/60.5 \\ \footnotesize{$\pm$ 0.4/0.1}} & \makecell*[c]{72.9/60.7 \\ \footnotesize{$\pm$ 0.3/0.2}} & \makecell*[c]{73.0/61.3 \\ \footnotesize{$\pm$ 0.2/0.2}} & \makecell*[c]{73.4/60.8 \\ $\pm$ \footnotesize{0.3/0.4}} & \makecell*[c]{\underline{74.1}/\underline{61.7} \\ \footnotesize{$\pm$ 0.3/0.2}} \\

\bottomrule
\end{tabular}
\caption{Cross-lingual dependency parsing results by language (UAS\%/LAS\%). We order the languages by order typology distances~\cite{ahmad-etal-2019-difficulties} to English. We use "$\downarrow$" to indicate results below mBERT and use \underline{underlined} text to highlight the best performance.}
\label{tab:main}
\end{table*}

\paragraph{Implement Details}

We employ mBERT~\cite{devlin-etal-2019-bert} to derive the cross-lingual word embeddings. Since the mBERT embeddings are subword-level, we follow previous work~\cite{wu-dredze-2019-beto, Liu_Winata_Cahyawijaya_Madotto_Lin_Fung_2021} in taking the first subword as word-level embeddings. The maximum subword sequence length is 512. We train the POS embeddings from scratch and set the dimension size of POS embeddings as 50. The batch size is set to 32. We use Adam~\cite{DBLP:journals/corr/KingmaB14} to train models with $\beta_1$ of 0.9, $\beta_2$ of 0.9 and $L2$ weight decay of 1e-5. The model is trained for 50 epochs with the learning rate of 1e-5 for mBERT and 3e-5 for other network layers. We choose the best hyper-parameters according to the development set in the source language. We empirically set $\lambda_1=1, \lambda_2=0.001$. We implement our method using PyTorch 1.8.0 based on \textit{Hugging Face} transformer library \footnote{https://huggingface.co}. 

To quantify the \textit{word order distance} between two languages, we following~\cite{ahmad-etal-2019-difficulties} select the 52 most frequently occurring dependency triples (Dep $\curvearrowleft$ Head, Label) across 31 languages, then concatenate relative frequency of the left-direction (dependent word before its head)  for all triples in each language as the word order feature. We use Manhattan distance as our word order distance. Due to differences in the versions of the datasets used, our word order distance does not strictly align with previous word order distance ~\cite{ahmad-etal-2019-difficulties} \footnote{Using order typology distance to distinguish it from ours}.

\paragraph{Baseline Parsers}
We compare our IWR-KD with several competing baselines involving word order learning as follows. \underline{\textit{SelfAtt-Direct}}~\cite{ahmad-etal-2019-difficulties} adopts the self-attention based  order-free encoder for cross-lingual parsing. \underline{\textit {mBERT-Direct}}~\cite{wu-dredze-2019-beto} is fine-tuned by adding the graph-based biaffine dependency parser on top of it. \underline{\textit{Frozen PE}}~\cite{Liu_Winata_Cahyawijaya_Madotto_Lin_Fung_2021} freezes the positional embedding in mBERT during the fine-tuning. \underline{\textit{Subtree-EWR}}~\cite{arviv-etal-2023-improving} explicitly reorders words through subtree-aware constraints for cross-lingual parsing. \underline{\textit{WOL}} is a variant of ours, which learns the word order knowledge in the source language with an auxiliary loss. For fair comparison, all methods are implemented with the same datasets and training configurations. We report the results of {\textit{SelfAtt-Direct}} in the original paper and re-implement other methods. Note that we do not compare ours with methods that use extra resources, e.g., \textit{SubDP}~\cite{shi-etal-2022-substructure} utilizes translation and word alignment to augment data, \textit{SFDP}~\cite{sun-tpami-sfdp} employs large amounts of unlabelled data to improve the performance.

\begin{table}[t]
    \centering
    % \label{tab:llm_com}
    \begin{tabular}{l | c }
        \toprule
        \textbf{Model} & \textbf{UAS\%/LAS\%} \\
        \midrule
        \textbf{IWR-KD (Ours)} & \textbf{74.1/61.7} \\ 
        w/ Pseudo-Labelling  & 73.5$_{-0.6}$/61.3$_{-0.4}$    \\ 
        w/ Silver POS Tags & 73.4$_{-0.7}$/60.9$_{-0.8}$    \\
        w/o FT & 72.6$_{-1.5}$/60.3$_{-1.4}$      \\
        \bottomrule
    \end{tabular}
    \caption{Ablation study (averaged results over all languages).}
\label{tab:main}
\end{table}

\begin{table*}[t]
    \centering
        \begin{tabular}{ ccc}
        \toprule
        \textbf{(Dep} $\curvearrowleft$ \textbf{Head, Label)}  & \textbf{Word Order Frequency} & \textbf{Source/Target Examples}\\
        \midrule
        (PRON $\curvearrowleft$ VERB, obj) & \makecell*[l]{\textbf{EN:} 0.1351 \\ \textbf{EN+IWR-KD:} 0.9348 \\ \textbf{FR:} 0.9109}  & \makecell*[l]{\textbf{EN:}Recently I 'm having trouble \colorbox{green!20}{training$^{\rm head}_{\rm VERB}$} \colorbox{red!20}{him$^{\rm dep}_{\rm PRON}$} . \\ \textbf{FR:} ... se sont permis de \colorbox{red!20}{les$^{\rm dep}_{\rm PRON}$} \colorbox{green!20}{traiter$^{\rm head}_{\rm VERB}$} en tant ...} \\
        \midrule
        (NOUN $\curvearrowleft$ VERB, obl) & \makecell*[l]{ \textbf{EN:} 0.0996 \\ \textbf{EN+IWR-KD:} 0.5055 \\ \textbf{DE:} 0.6031} & \makecell*[l]{ \textbf{EN:} ... be \colorbox{green!20}{causing$_{\rm VERB}^{\rm head}$} us trouble for \colorbox{red!20}{years$_{\rm NOUN}^{\rm dep}$} to come. \\ \textbf{DE:} ... französischen \colorbox{red!20}{Radsportverband$_{\rm NOUN}^{\rm dep}$} \colorbox{green!20}{gezahlt$_{\rm VERB}^{\rm head}$} ... } \\ 
        \midrule
        (NOUN $\curvearrowleft$ NOUN, nmod) & \makecell*[l]{ \textbf{EN:} 0.0056 \\ \textbf{EN+IWR-KD:} 0.4867 \\ \textbf{ET:} 0.8296} & \makecell*[l]{ \textbf{EN:} Hope there is useful \colorbox{green!20}{grist$^{\rm head}_{\rm NOUN}$} for the \colorbox{red!20}{mill$^{\rm dep}_{\rm NOUN}$} here . \\ \textbf{ET:} ...  \colorbox{red!20}{tänapäeva$^{\rm dep}_{\rm NOUN}$} ühe olulisema ameerika \colorbox{green!20}{luuletaja$^{\rm head}_{\rm NOUN}$} } \\ 
        \bottomrule
        \end{tabular}
\caption{Case study on Implicit Word Reordering. Word order frequency indicates the relative frequency of the left direction (dependent before its head). The GREEN (RED) highlight indicates a head (dependent) word.}
\label{tab:case}
\end{table*}

\subsection{Main Results}
Table \ref{tab:main} presents the results on the test sets. The languages are ordered by their order typology distance to English. From the experimental results, we can make the following observations. 
(1) Our proposed IWR-KD method achieves state-of-the-art performance on most languages and average performance across all languages, demonstrating the effectiveness and generalization of IWR-KD.
(2) The order-sensitive direct transfer method (mBERT) significantly surpasses the traditional order-agnostic self-attention method (SelfAttn) by effectively capturing richer contextual information, including word order.
(3) The performance of the mBERT-based word order-agnostic model (Frozen PE) declines relative to direct transfer (mBERT) in both the source language and similar languages, such as English (en) and Norwegian (no), due to underfitting caused by the frozen position representation.
(4) The explicit word reordering method (Subtree-EWR) substantially declines in some languages, such as Slovenian (sl) and Finnish (fi), due to unnatural sentence rearrangements that introduce noise detrimental to model learning. Note that the EWR method indirectly accesses the target language dependency annotation information~\cite{arviv-etal-2023-improving}, so it can achieve high performance in some languages. While IWR-KD relies solely on low-cost target language POS tags, it outperforms SubtreeWR in average performance.
(5) WOL performs better in languages closer to English, while mBERT outperforms it in languages that are more "distant" from English. This indicates capturing word order is critical in cross-lingual dependency parsing. 

\subsection{Ablation Study}
To verify the effect of each part of our method, we do experiments with the following variants of our IWR-KD. 
(1) \textit{w/ pseudo-labelling}, which converts the soft word order probability distribution into hard labels to guide the student model training. Hard labels result in performance degradation, highlighting that our word order distillation loss conveys richer word order knowledge than hard labels.
(2) \textit{w/ Silver UPOS}, which employs the Stanza \footnote{https://stanfordnlp.github.io/stanza} tool to annotate POS tags to simulate a more realistic application scenario. In this case, our method can improve the performance of the direct transfer method (mBERT in Table \ref{tab:main}).
(3) \textit{w/o FT}, which eliminates the fine-tuning of mPLMs to make the model more lightweight. However, performance drops significantly without fine-tuning, indicating that fine-tuning enables the encoder to effectively adapt to the specific task.

\subsection{Case Study}

In this section, we conduct a series of case studies to bring up insights into why the proposed IWR-KD works. 

The IWR-KD method can help the parser correctly identify dependency labels by exploiting the learned order relationship between dependent words and their heads. Specifically, if the model can align the word order relationships of the source language to the target language, misidentified dependency labels in the target language may be corrected. As shown in Table~\ref{tab:case}, in the first example (PRON $\curvearrowleft$ VERB, obj), the relative frequency of PRON before VERB differs greatly between the source and target languages, which leads to migration failure. Our model's prediction of the relative frequency of PRON before VERB in the source language is close to that of the target language. At the same time, our model learned that their label is obj on the source language training set, and then the model can correctly predict PRON and PRON in the target language. The dependency tag of VERB is obj. Two additional examples present the same results using different languages.

\begin{figure*}[!t]
\centering
\begin{tabular}{c}
     \includegraphics[width=0.99\linewidth]{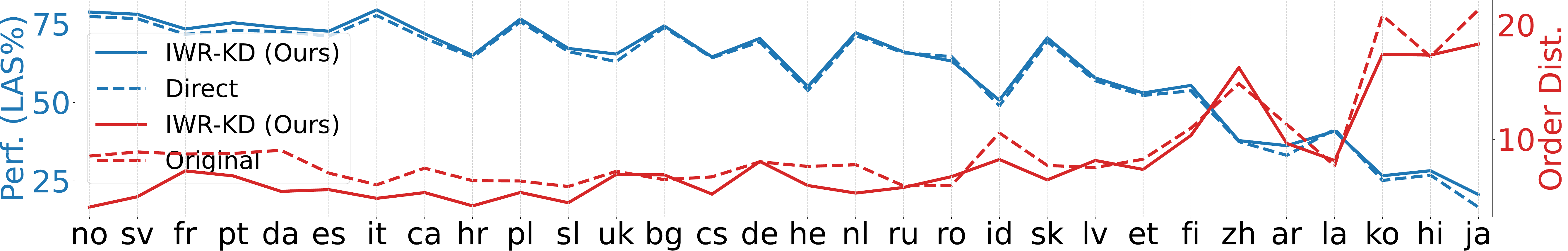}  \\ 
     \includegraphics[width=0.99\linewidth]{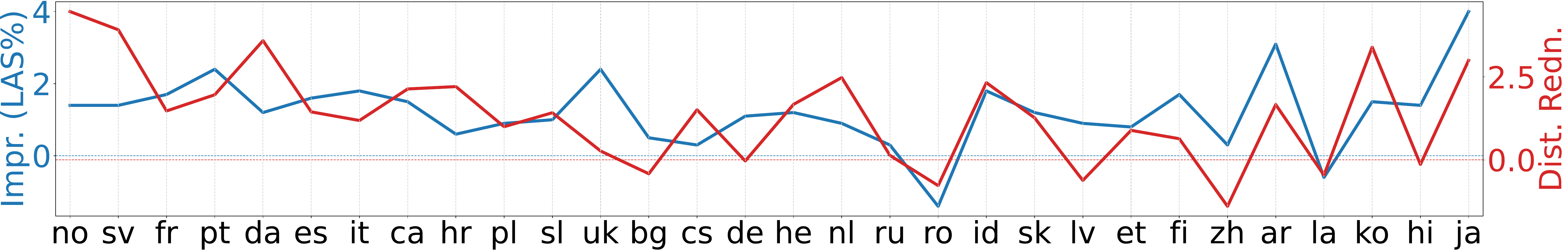}    \\
\end{tabular}
\caption{Word order distance and performance. Languages (x-axis) are sorted by their order typology distances~\cite{ahmad-etal-2019-difficulties} to English from left to right.}
\label{exp:performance_wo_distance}
\end{figure*}

\begin{figure}[!t]
\centering
\begin{tabular}{ccc}
    \hspace{-0.35cm} \includegraphics[width=0.32\linewidth]{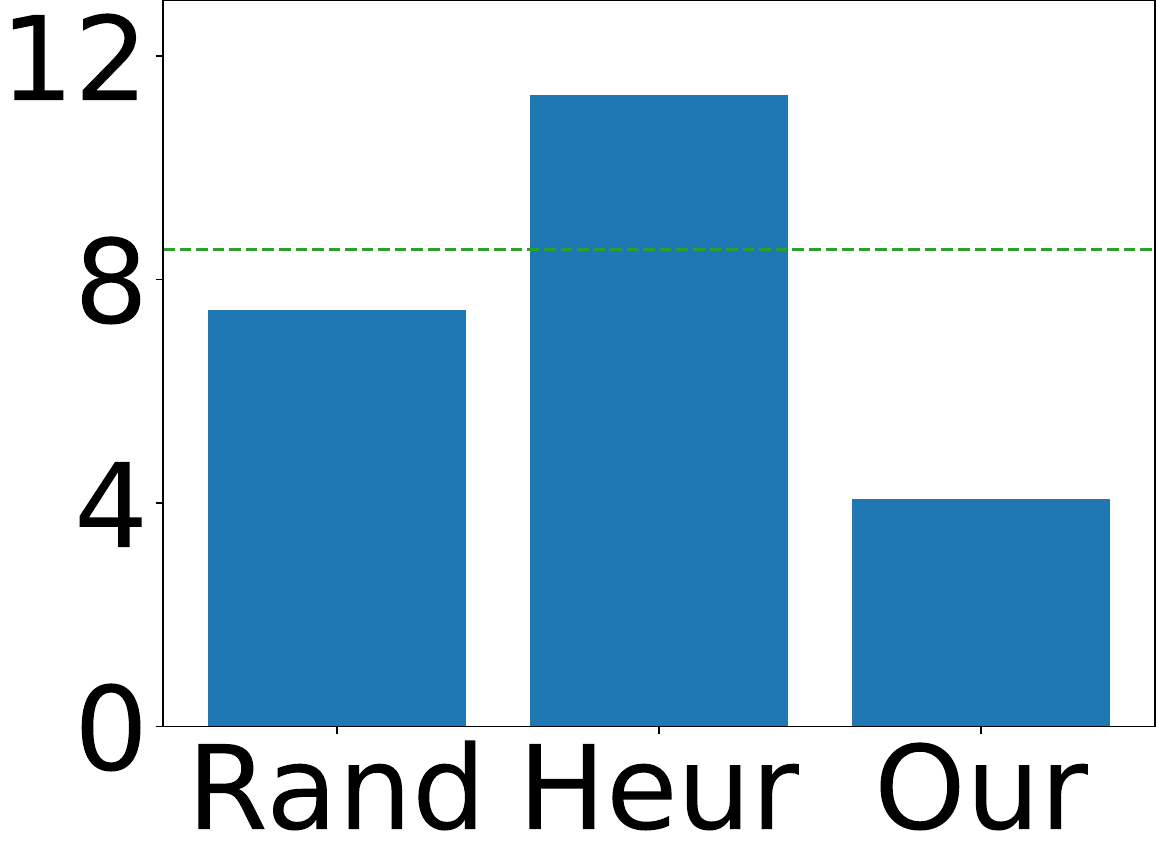} &
    \hspace{-0.35cm} \includegraphics[width=0.32\linewidth]{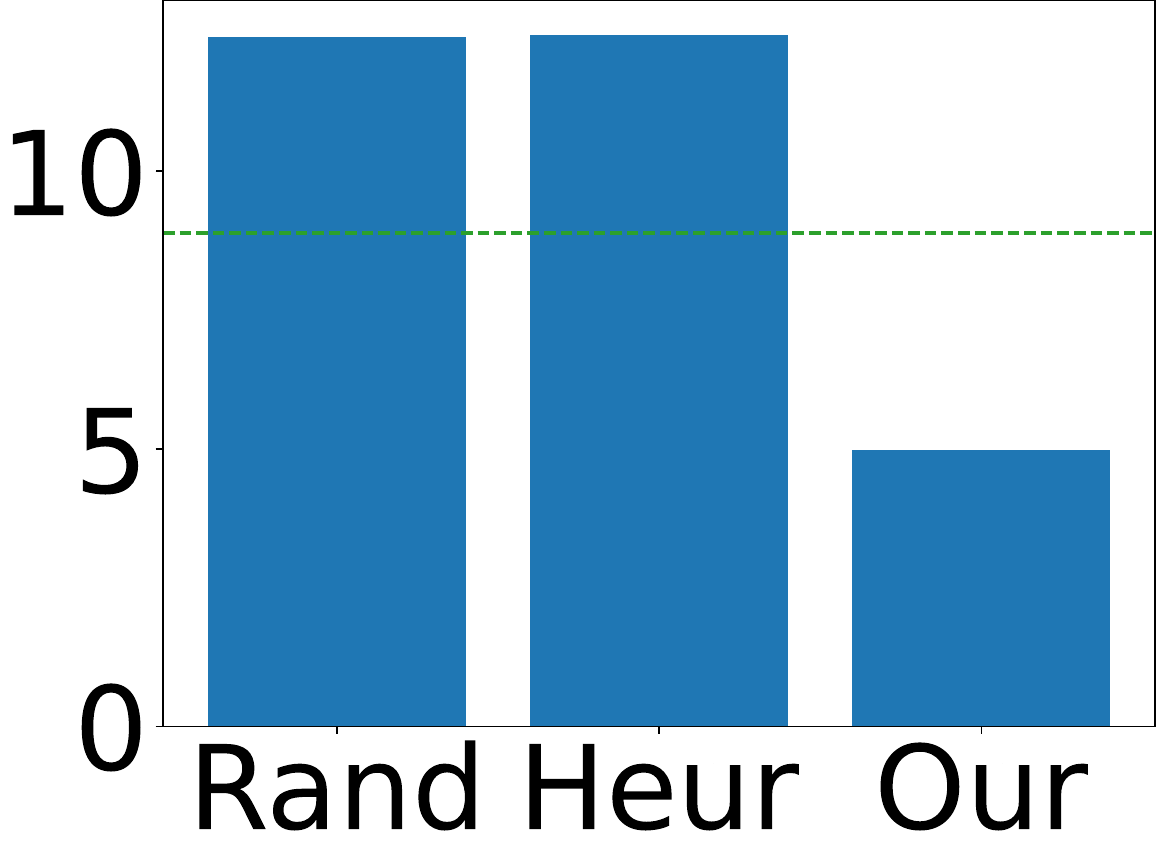} &  \hspace{-0.35cm} \includegraphics[width=0.32\linewidth]{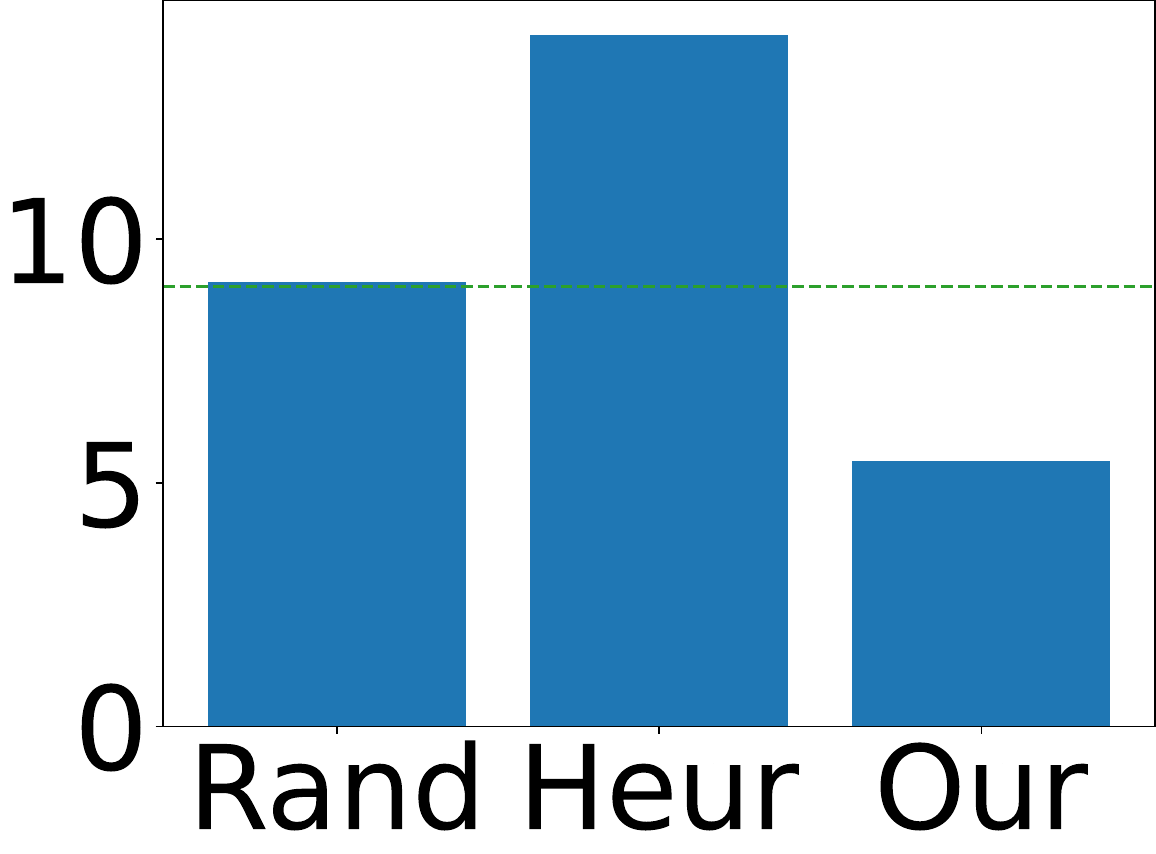} \\
    \hspace{-0.35cm} no &
    \hspace{-0.35cm} sv &  
    \hspace{-0.35cm} da \\
    \hspace{-0.35cm} \includegraphics[width=0.32\linewidth]{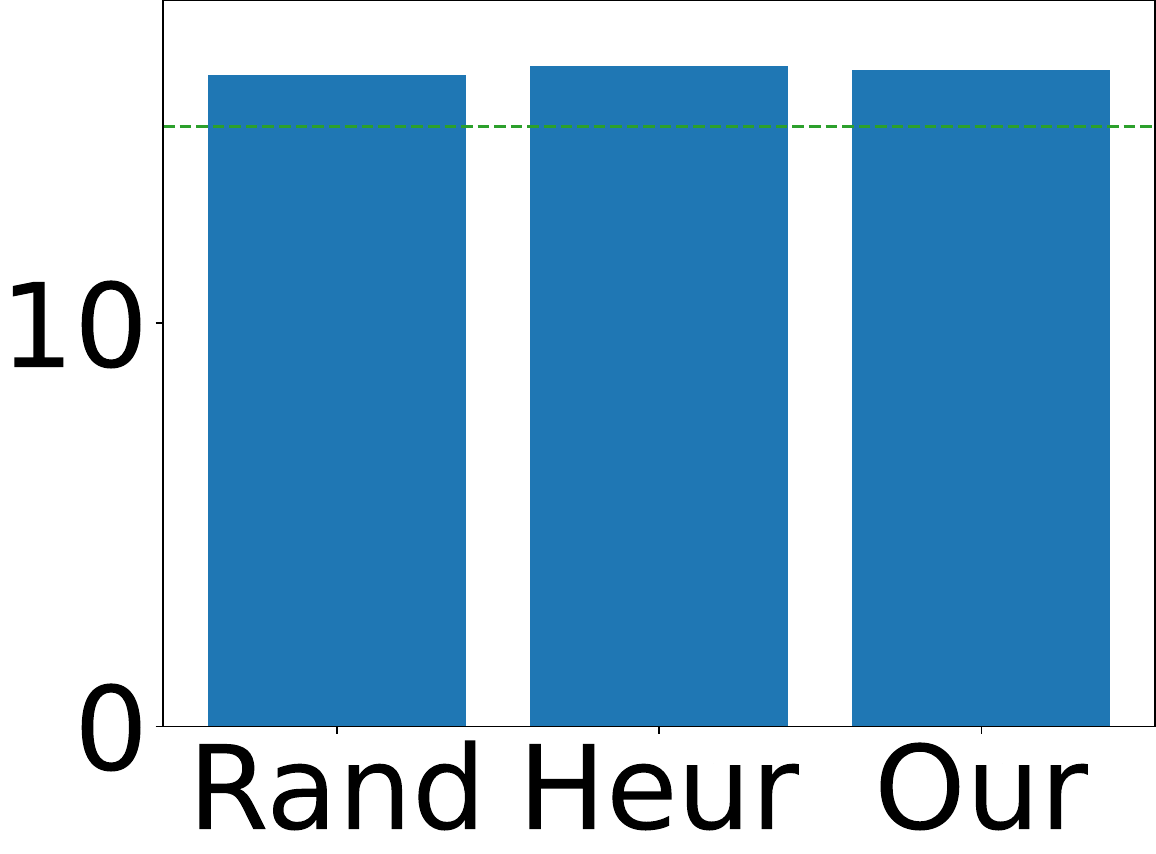} &
    \hspace{-0.35cm} \includegraphics[width=0.32\linewidth]{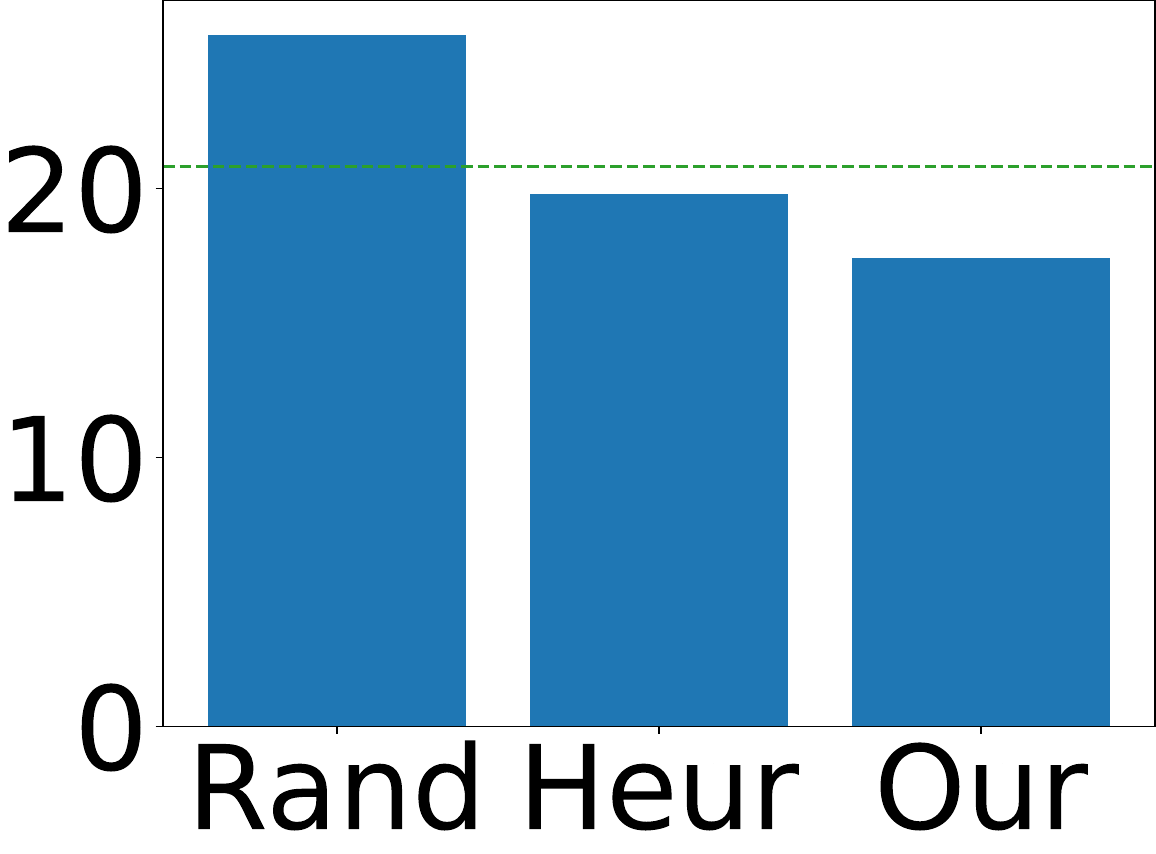} &  \hspace{-0.35cm} \includegraphics[width=0.32\linewidth]{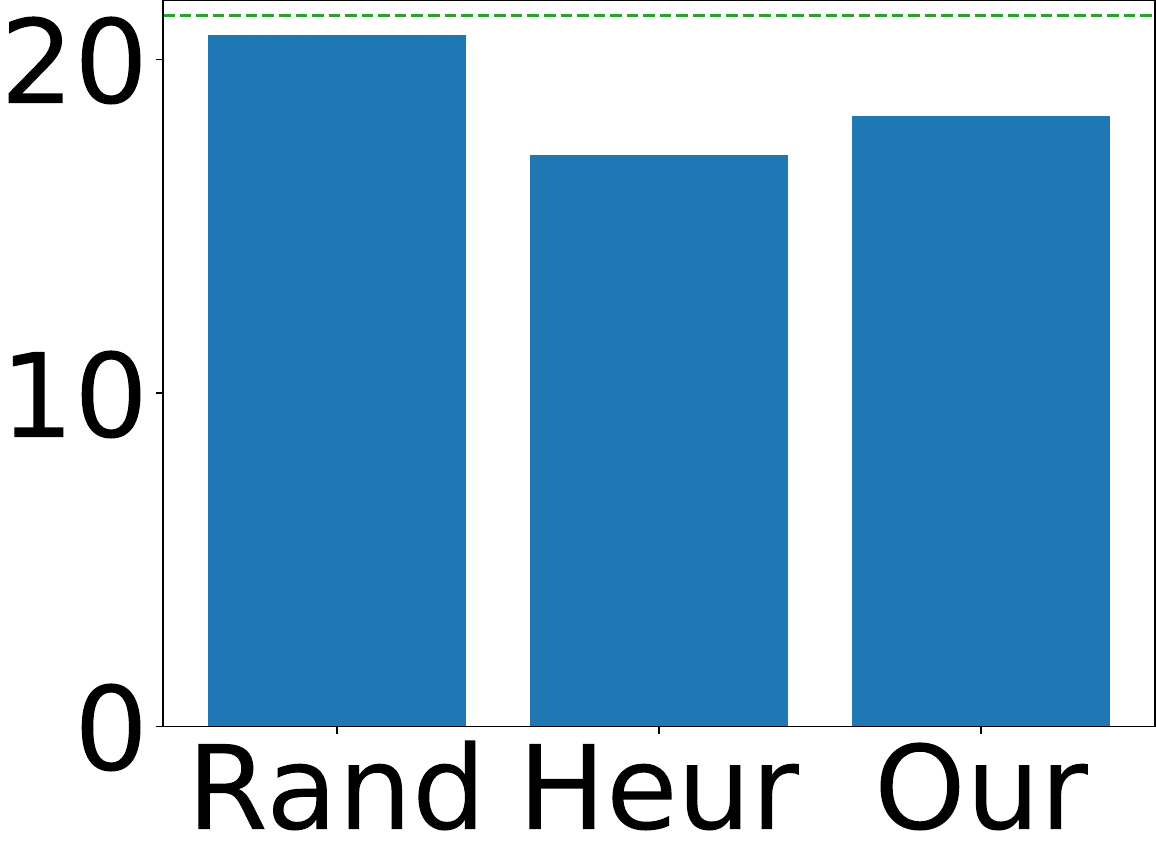} \\
    \hspace{-0.35cm} zh &
    \hspace{-0.35cm} ko &  
    \hspace{-0.35cm} ja \\
\end{tabular}
\caption{Word order distance predicted by different word reordering teachers. The green bar indicates original word order distance between English and the target language.}
\label{exp:wo_teachers}
\end{figure}

\subsection{Performance versus Word Order Distance}

As shown in the upper subfigure of Figure \ref{exp:performance_wo_distance}, transfer ability decreases as the word order distance increases from left to right, indicating that word order is a crucial factor for transfer performance (Perf.). The Pearson correlation coefficient between word order distance and transfer performance is -0.8803 (p-value = 1e-10), demonstrating a strong negative correlation. As shown in the lower subfigure of Figure \ref{exp:performance_wo_distance}, the reduction (Redn.) in word order distance positively influences performance improvement (Impr.). The Pearson correlation coefficient is 0.4308 (p-value = 0.0175), indicating a strong positive correlation.

\subsection{Effect of Different Teacher Models}
This section investigates the effect of the different word reordering teacher models. We compare three learning methods: (1) \textit{Rand:} Randomly learning the word order relationship between pairs of words. (2) \textit{Heur:} Heuristically learning the ordering relationships of pairs belonging to 52 selected dependency triples. (3) \textit{Our:} Based on a pre-trained parser in the source language, we predict the head of each word and learn their ordering relationships. 

As shown in Figure \ref{exp:wo_teachers}, we calculate the word order distance predicted by three different teacher models on six different target languages. Three languages, i.e. no, sv, da, are closer to the source language, while the other three languages, i.e. zh, ko, ja, are far away from the source language. It can be seen that our transfer-based teacher model can effectively reduce the word order distance between the source language and target languages. We notice that our method fails to reduce the word order distance between the source language and zh (Chinese), but it improves the transfer performance as in Table \ref{tab:main}. 
Through more thorough observation, we find that our method reduces the word order difference on some dependency triples. For example, for the word order frequency of (NOUN $\curvearrowleft$ NOUN, nmod), EN: 0.0056, ZH: 0.9980, EN+IWR-KD: 0.5413. We conjecture that different dependencies have different effects on transfer performance, which may be one of our future research directions.

\section{Related Work}

Word order refers to the arrangement of words in a sentence or phrase to convey meaning in a particular language. In the cross-lingual community, word order is widely explored from two perspectives: order-agnostic methods~\cite{ahmad-etal-2019-difficulties, Liu_Winata_Cahyawijaya_Madotto_Lin_Fung_2021, ding-etal-2020-self, DBLP:journals/corr/abs-2305-19857, DBLP:journals/corr/abs-1910-12391, hessel-schofield-2021-effective} and word reordering methods~\cite{rasooli-collins-2019-low, ji-etal-2021-word, liu-etal-2020-cross-lingual-dependency, chen-etal-2019-neural, goyal-durrett-2020-neural, al-negheimish-etal-2023-towards, pham-etal-2021-order}. The former argues that word order encoding is a risk for cross-lingual transfer, as the model often fits in the language-specific order. Reducing the word order information fitted into the models can improve the cross-lingual adaptation performance in position representation~\cite{ding-etal-2020-self} and dependency parsing~\cite{ahmad-etal-2019-difficulties, Liu_Winata_Cahyawijaya_Madotto_Lin_Fung_2021}. The latter is dedicated to reordering the word from one language to another. For example, Arviv et al.~\cite{arviv-etal-2023-improving} achieves better cross-lingual transfer results by rearranging the words in the source language to meet the word order in the target language conditioned on the syntactic constraints. 

\section{Conclusion}
In this paper, we explore a new attempt, \textit{implicit word reordering}, and propose an implicit word reordering method with knowledge distillation (\textbf{IWR-KD}), which uses the word reordering model as a teacher to guide the student model to simultaneously learn target-language word order knowledge and source-language dependency parsing knowledge on the labeled data of the source language. IWR-KD can effectively capture vital word order information and does not require the actual generation and selection of reordered sentences, which addresses the limitations of previous order-agnostic and explicit word reordering methods. We conduct extensive experiments on 31 different languages on Universal Dependency Treebanks, which showed that IWR-KD outperforms multiple competitive methods.

\section*{Acknowledgements}
We thank Beijing Advanced Innovation Center for Big Data and Brain Computing for providing computation resources. We also thank the anonymous reviewers and the area chair for their insightful comments. This work was supported by the National Natural Science Foundation of China (No. U2433212, No. 62306026), in part by the China Postdoctoral Science Foundation (No. 2023M740184), in part by the Fundamental Research Funds for the Central Universities, and in part by the State Key Laboratory of Complex \& Critical Software Environment.

\end{document}